\documentclass[fleqn,10pt]{wlscirep}
\usepackage[utf8]{inputenc}
\usepackage[T1]{fontenc}
\usepackage{amsmath} 
\usepackage{amssymb}
\usepackage{subfigure}
\usepackage{cite}
\usepackage{hyperref} %放在所有宏包最后
\usepackage{nameref}

\title{Hybrid Gaussian Process Regression with Temporal Feature Extraction for Partially Interpretable Remaining Useful Life Interval Prediction in Aeroengine Prognostics}

\author[1]{Tian Niu}
\author[1]{Zijun Xu}
\author[1]{Heng Luo}
\author[1,*]{Ziqing Zhou}
\affil[1]{Fudan University, 220 Handan Road, Shanghai, China}
\affil[*]{Email: ziqingzhou21@m.fudan.edu.cn}

\begin{abstract}
The estimation of Remaining Useful Life (RUL) plays a pivotal role in intelligent manufacturing systems and Industry 4.0 technologies. While recent advancements have improved RUL prediction, many models still face interpretability and compelling uncertainty modeling challenges. This paper introduces a modified Gaussian Process Regression (GPR) model for RUL interval prediction, tailored for the complexities of manufacturing process development. The modified GPR predicts confidence intervals by learning from historical data and addresses uncertainty modeling in a more structured way. The approach effectively captures intricate time-series patterns and dynamic behaviors inherent in modern manufacturing systems by coupling GPR with deep adaptive learning-enhanced AI process models. Moreover, the model evaluates feature significance to ensure more transparent decision-making, which is crucial for optimizing manufacturing processes. This comprehensive approach supports more accurate RUL predictions and provides transparent, interpretable insights into uncertainty, contributing to robust process development and management.
\end{abstract}
\begin{document}

\flushbottom
\maketitle
\thispagestyle{empty}

\section*{Introduction}

Prognostics and Health Management (PHM) plays a crucial role in increasing equipment availability, reducing maintenance costs, and improving the scheduling of maintenance events. By predicting potential failures, maintenance activities can be planned, reducing downtime and allowing for more efficient operations in manufacturing environments~\cite{das2011essential}. Predicting a machine’s Remaining Useful Life (RUL) using real-time condition data is central to effective prognostics~\cite{lei2018machinery}. With modern machines continuously collecting sensor data at frequent intervals, this data forms a time series with inherent temporal characteristics. Consequently, RUL prediction involves solving a time series regression problem that captures both the operational and mechanical characteristics of the machinery~\cite{ferreira2022remaining}.

Like many other time series prediction challenges, RUL estimation methods can be divided into two main categories: learning-based and model-based approaches. Learning-based methods utilize machine learning and deep learning techniques, such as long short-term memory~\cite{6795963} and transformer~\cite{vaswani2017attention},
which leverage historical data to improve predictions. These methods rely on supervised learning to achieve accurate outcomes. In manufacturing processes, it is critical to incorporate temporal continuity and system behavior dynamics when dealing with high-dimensional data. However, many traditional machine learning methods focus solely on single-point predictions without quantifying uncertainty, limiting their reliability in real-world applications.

Model-based approaches, on the other hand, include both physical and statistical methods. Physical approaches involve building models that represent the underlying mechanical systems. Implementing this approach necessitates intricate physical modeling, and the inherent variability among different machines poses challenges to its generalizability, thereby constraining its feasibility for broader, real-world applications~\cite{chan2012life}. In contrast, statistical methods, particularly Gaussian Process Regression (GPR), offer a more flexible way of predicting RUL. GPR is a non-parametric Bayesian approach that effectively models diverse deterioration mechanisms and outputs uncertainty measures, making it highly applicable in complex systems where physical models may fail~\cite{schulz2018tutorial,liu2019remaining}. Despite its robustness in generating prediction intervals, GPR alone may face challenges in handling large-scale time-series data typical of industrial environments. Specifically, GPR involves calculating and storing an $n\times n$ covariance matrix (where $n$ is the number of data points), resulting in a computational complexity of $O(n^3)$ and a memory complexity of $O(n^2)$.

This paper proposes a novel Hybrid Gaussian process Regression with temporal feature extraction for partially interpretable RUL interval Prediction in aero-engine prognostics (HRP) to overcome the limitations of both learning and model-based approaches. The approach leverages a modified GPR model, pre-trained on time-series data, to incorporate temporal dynamics and system behaviors relevant to intelligent manufacturing systems. Unlike single learning-based methods, this method captures uncertainty and key predictive features without relying solely on black-box neural networks by modifying GPR into deep adaptive learning-enhanced regression. The pre-processing steps ensure that the modified GPR effectively utilizes feature representation and system diagnostics to increase robustness compared to single model-based methods. Additionally, the model identifies and prioritizes critical physical features contributing to system degradation, enhancing predictive accuracy and enabling optimal maintenance decisions.

The key contributions of this paper are as follows:

\begin{itemize}
\item[$\bullet$] A supervised GPR model pre-trained with temporal features is proposed to reflect system health states in manufacturing environments. This model is specifically tailored to capture the time-dependent degradation of machinery, enhancing predictive insights into system performance over time.

\item[$\bullet$]Confidence intervals generated by the GPR model quantify the uncertainty of predicted RUL, offering a dual advantage: improved prediction accuracy and a more precise assessment of failure risk.
\item[$\bullet$] Feature importance analysis is incorporated to identify key contributors from mechanical and sensor data, providing actionable insights into degradation factors. The model facilitates real-time, data-driven maintenance planning by integrating uncertainty modeling and adaptive learning techniques.
\end{itemize}

The rest of this paper is organized as follows: First, we introduce the preliminaries and outline the proposed method. Then, we discuss the experimental results and analysis. Finally, we summarize the findings and discuss future directions for RUL estimation in smart manufacturing environments.

\section*{Related works}

RUL prediction has been a critical area of research in prognostics and health management for enhancing equipment availability and reducing downtime. While various methods have been proposed, most approaches face challenges in balancing predictive accuracy with interpretability and uncertainty quantification. 
Historically, learning-based methods, such as long short-term memory networks and Transformer models, have been widely used for RUL prediction due to their ability to capture temporal dynamics in sequential data~\cite{nelson2017stock, altche2017lstm}. Liu et al.~\cite{liu2021prediction} integrated clustering analysis with LSTM to enhance prediction accuracy. In contrast, Wang et al.~\cite{wang2021adaptive} combined adaptive sliding windows with LSTM to predict the performance of lithium-ion batteries. However, these methods focus on point predictions and often lack interpretability, especially regarding feature significance and uncertainty quantification. Additionally, while LSTM models effectively capture sequential patterns, they frequently function as black-box models, providing little insight into the underlying mechanics of degradation processes.

Researchers employ statistical methods like GPR for RUL prediction to address these limitations, offering a more interpretable and flexible approach. GPR, as a non-parametric Bayesian technique, provides both point estimates and confidence intervals, thus quantifying uncertainty in predictions~\cite{schulz2018tutorial,liu2019remaining}. This makes GPR particularly valuable in scenarios where the complexity of machinery precludes purely physical modeling~\cite{chan2012life}. For example, Hong et al.~\cite{hong2012remaining} applied GPR to predict bearing RUL by modeling the relationship between time-domain features and future operational states. Similarly, Baraldi et al.~\cite{baraldi2015prognostics} demonstrated the utility of GPR in modeling creep growth in materials, providing reliable prediction intervals for degradation processes.

However, despite its advantages in uncertainty quantification, GPR faces challenges in processing large-scale time-series data, which is typical in industrial applications. Previous studies have shown that GPR's scalability can be a limitation when handling high-dimensional datasets. Some researchers have explored hybrid approaches that integrate GPR with other learning techniques to mitigate this. For instance, Pang et al.~\cite{pang2022interval} employed fuzzy information granulation with least squares support vector machines to enhance time interval forecasting for lithium-ion batteries. However, such methods still rely heavily on feature engineering and lack adaptive learning capabilities.

Our method enhances GPR by incorporating temporal dynamics through adaptive learning, enabling it to handle large datasets and improve uncertainty modeling. This approach bridges interpretability, uncertainty quantification, and predictive accuracy for RUL estimation, making it well-suited for Industry 4.0 applications. In summary, this work combines machine learning and statistical modeling to provide a transparent, robust solution for real-time, data-driven maintenance in industrial settings.

\section*{Methods}
\phantomsection
\label{section3}

This section begins by formulating the multi-dimensional time series problem in RUL prediction. Then, an overview of our interval prediction framework is provided. The model is referred to using the acronym HRP. After that, we introduce the details of the model's main elements. 

\subsection*{Problem definition}

Given run-to-failure data for several mechanical systems, denoted as $\left \{ \mathbb{X} _{t}  \right \}_{t=1}^{T}  $  and corresponding RUL: $\mathbf{y}=\left \{ y_{t}  \right \}_{t=1}^{T}  $, where $\mathbb{X} _{t}  = \left \{ x_{t}^{1} , x_{t}^{2} ,...,x_{t}^{M} \right \} $ represents observations at the corresponding running time t,  $M$ represents the number of monitoring sensors, and $T$ represents the total operational life of a machine. Consequently, the  RUL of time $t$ is defined as $y_{t}=T- t$. The objective is to find a mapping from high-dimensional observations to a scalar value, defined as $f:\left \{ \mathbb{X} _{t}  \right \}_{t=1}^{T}\mapsto \left \{ y _{t}  \right \}_{t=1}^{T}$. Given new data $\left \{ \mathbb{X} _{t}  \right \}_{t=1}^{K}  $, estimates of  $ \left \{y _{t}  \right \}_{t=1}^{K}$ can be obtained as $ \left \{ \hat{y}  _{t}  \right \}_{t=1}^{K}=f\left (  \left \{ \mathbb{X}  _{t}  \right \}_{t=1}^{K} \right ) $, and predict the confidence interval$ \left \{ \left [   y ^{L} _{t} , y ^{U} _{t}\right ]\right \}_{t=1}^{K}$. Simultaneously, the influence of the $M$ monitoring sensors on faults is analyzed, enabling the estimation of sensor impact factors  $\left \{ \lambda  ^{1} ,\lambda ^{2} ,...,\lambda ^{M} \right \} $. 

\subsection*{Framework overview}

\begin{figure}
    \centering
    \includegraphics[width=\textwidth]{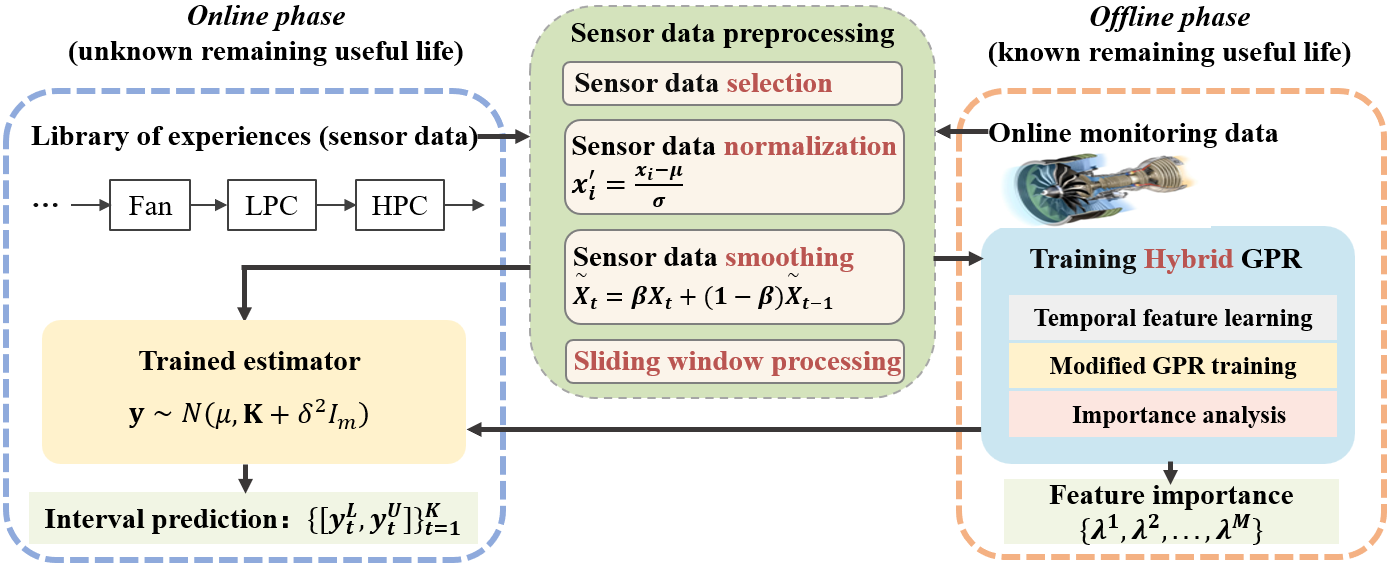}
    \caption{Dual-line interval RUL prediction framework.}
    \label{framework_1}
\end{figure}

The dual-line interval RUL prediction strategy, illustrated in Fig.\ref{framework_1}, consists of two key stages: offline training and online prediction. This strategy provides engineers with two crucial outputs: the sensor influence on fault modes and the predicted interval for the RUL. Initially, all data follow the same preprocessing pipeline, including sensor selection, normalization, smoothing, and sliding window process. During training, the network is trained with data from known run-to-failure cycles, aligning the real RUL with a health index. The trained model serves both as a callable real-time prediction tool and a mechanism to provide insights into the relative importance of features during fault cycles. In the test phase, incomplete degradation trajectories from test data are processed in real time, providing rapid predictions using the trained network. 

\begin{figure}
    \centering
    \includegraphics[width=\textwidth]{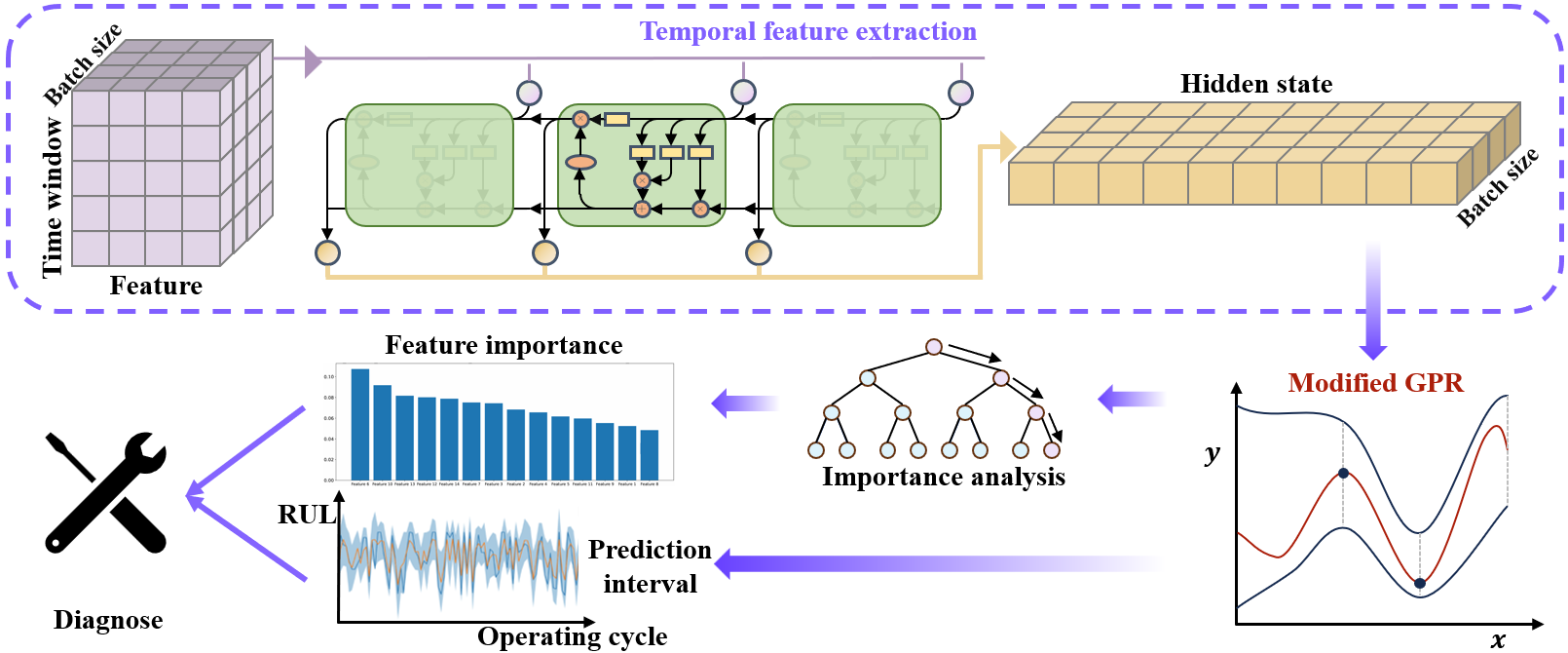}
    \caption{Illustration of the network in our method HRP.}
    \label{framework_2}
\end{figure}

The detailed structure of this network is depicted in Fig.~\ref{framework_2}, and follows a series of steps. Firstly, after data have been preprocessed, raw data are converted into the format 
$\left ( Batch~size
, Time~window~length, Feature~number\right ) $
, which serves as the input to the network. Secondly, temporal feature extraction adopts mathematical techniques for managing temporal information flow. Adaptive filtering and time-series compression gated memory cells preserve long-term dependencies, ensuring that relevant historical data are retained in the feature set while discarding noise or less significant information. This preprocessing step compresses the sequence of high-dimensional sensor data into a lower-dimensional latent space with a dimension of $\left ( Batch~size, Hidden~state~length\right ) $. 
The extracted time-series features are then fed to construct mean function $\mu$ and the covariance function $\mathbf{K}$ in multivariate normal Gaussian distribution, which not only learns the distribution of the data but also generates prediction intervals for RUL with quantified uncertainty.
The model's ability to capture temporal dynamics and represent the system's health status probabilistically is key to its enhanced performance. Furthermore, the network performs feature importance analysis based on how different sensors contribute to the fault modes, providing engineers with a clearer understanding of the degradation process.
In the final step, engineers can conduct a comprehensive run-to-failure assessment based on the prediction intervals produced by modified GPR and the detailed feature analysis. This process allows for both predictive insights and interpretability, combining the power of advanced temporal modeling with the flexibility and uncertainty quantification of modified GPR.

\subsection*{Data preprocessing}

We employ a four-step procedure to process the original dataset to ensure the data is prepared for practical prognostics analysis. First, we perform feature selection. After an extensive review of the sensor signals, 14 critical sensors are retained, providing the most significant information for accurate RUL predictions. Additionally, we utilize a piecewise linear degradation model to model the RUL as described in previous research~\cite{shi2024dual}. Here, the time for the change point is set at 125. 
Second, we apply z-score normalization to the sensor signals for each instance, standardizing the readings to improve model performance and comparability across different sensors~\cite{yu2019remaining}. This normalization transforms the original time series $x_{i}^{j}$, corresponding to the $j$-th sensor of the $i$-th sample, as formula $x_{i}^{j\prime}=\frac{x_{i}^{j}-\mu^{j}}{\sigma^{j}}$ ,  
where $\mu^{j}$ and $\sigma^{j}$ represent the mean value and the standard deviation deviation, respectively, for the $j$-th sensor readings across all instances in each dataset. Third, we implement exponential smoothing~\cite{shi2024dual} to retain the underlying trends in sensor signals while minimizing the impact of noise and short-term fluctuations. This technique  improves the accuracy of predictive models by focusing on long-term degradation trends. Finally, we segment the preprocessed data using a sliding window approach~\cite{hota2017time}. This method partitions the entire time series into smaller, equal-length windows. The window length is determined based on prior empirical studies~\cite{liu2022aircraft}, ensuring it captures sufficient temporal information for reliable prediction. Detailed descriptions of these four preprocessing steps, including parameters and justifications, are provided in the Supplementary Materials under S2.

\subsection*{Temporal feature extraction}

Advanced temporal feature extraction methods are employed during pre-analysis to enhance the model's ability to capture temporal dependencies. These methods are based on the mathematical principles of sequential learning, which allow the model to retain and process long-term and short-term dependencies inherent in time-series data. The method utilizes gating mechanisms that control the flow of information through the model, mimicking how LSTM networks handle temporal data through input control, memory retention, and output management~\cite{hochreiter1997long,cheng2022dual}. Detailed mathematical expressions are provided in the Supplementary Materials under S3. Once the temporal features from input $\left \{ \mathbb{X} _{t}  \right \}_{t=1}^{L} =  \left \{ x_{t}^{1} , x_{t}^{2} ,...,x_{t}^{M} \right \}_{t=1}^{L}$ have been extracted, they are formatted into hidden states $\mathbf{h} =(\mathbf{h} ^{1} ,\mathbf{h} ^{2},...\mathbf{h} ^{m})$, which encapsulate the most important information from the time-series. These hidden states, which compress the multivariate time series into a reduced form, are then transferred to the next processing stage, modified GPR, where distribution learning occurs, and feature importance analysis is performed. 
The hidden states can be expressed as
\begin{eqnarray}
(\mathbf{h} ^{1} ,\mathbf{h} ^{2},...\mathbf{h} ^{m})=f_{\mathrm{tem}}(\left \{ \mathbb{X} _{t}  \right \}_{t=1}^{L};\mathbf{\theta}_{tem}).
\label{eq:hidden state}  
\end{eqnarray}
Here, $\theta_{tem}=\begin{bmatrix}\mathbf{W}_{tem},\mathbf{b}_{tem}\end{bmatrix}$ represents the learnable parameters of the feature extraction process and $m$ stands for the hidden layer size.

The loss function used to optimize the feature extraction process is designed to handle outliers in the data by employing a Huber loss, which combines the benefits of squared error and absolute error, defined as: 
\begin{eqnarray}
 L_\delta(a)=\begin{cases}
  & \frac12(a)^2~~~~~~~~~~~~~~~~~\text{ if } |a|\leq\delta \\
  & delta(|a|-\frac12\delta)~\text{ if } |a|> \delta
\end{cases}.
\end{eqnarray}

In this way, the extracted features retain the sequence's temporal dynamics while ensuring stability in the training process. They are also tailored to the requirements of the Gaussian Process framework for further learning. 

\subsection*{Modified GPR}

Different from classic GPR, we modify GPR cooperated with temporal learning to adaptively describe uncertainty. Modify GPR
 receives the hidden state $\mathbf{h} =(\mathbf{h} ^{1} ,\mathbf{h} ^{2},...\mathbf{h} ^{m})$ as defined by equation (\ref{eq:hidden state}), and produces outputs $\left \{ \hat{y}  _{t}  \right \}_{1}^{K}$ and confidence interval $ \left \{ \left [   y ^{L} _{t} , y ^{U} _{t}\right ]\right \}_{t=1}^{K}$. A Gaussian process (GP) is a collection of random variables, any finite subset that adheres to a joint Gaussian distribution. For any input $\mathbf{h}$, the probability distribution of the corresponding RUL ${\mathbf{y}} =\left \{ y_{t}  \right \}_{t=1}^{T}  $  follows a multivariate normal Gaussian distribution. The GP is characterized by two essential components, the mean function $\mu$ and the covariance function $\mathbf{K}$, which jointly specify its probability distribution,
\begin{eqnarray}
\mathbf{y}  \sim GP(\mathbf{\mu}, \mathbf{K}). \label{eq:quadratic7}
\end{eqnarray}
Generally, the mean function $\mu(\mathbf{h})$ is selected as a zero-mean function, and the covariance function $ k(\mathbf{h}, \mathbf{h}^{\prime})$ is the squared exponential function, defined as 
\begin{eqnarray}
     k(\mathbf{h}, \mathbf{h}^{\prime}) = \tau^2 \exp\left(-\frac{\|\mathbf{h} - \mathbf{h}^{\prime}\|^2}{2\eta^2}\right),\label{eq:quadratic8}
\end{eqnarray}
where $\eta$ represents the characteristic length-scale, and $\tau$ represents the amplitude of the covariance. Therefore, the mean vector $\mathbf{\mu}$ satisfies $\mathbf{\mu} = {(\mu(\mathbf{h}^{1}), \ldots, \mu(\mathbf{h}^{m}))}^{\intercal}$, and the $n$-by-$n$ covariance matrix $\mathbf{K}$ satisfies $\mathbf{K} = (k(\mathbf{h}^{i}, \mathbf{h}^{i'}))_{i, i'=1}^{m}$. Similar to GPR, our method HRP assumes that the observations are noisy realizations of the GP prior. The noise is typically modeled as a Gaussian distribution with zero mean and variance $\varepsilon_{m} \sim \mathcal{N}(0, \delta^{2})$. Given these assumptions, the prior distribution follows a multivariate normal distribution, where $I_m$ denotes the $m$-by-$m$ identity matrix, 
\begin{eqnarray}
\mathbf{y}\sim N\begin{pmatrix}\mathbf{\mu},\mathbf{K}+\delta^2I_m\end{pmatrix}.\label{eq:quadratic9}
\end{eqnarray}

The core strength of GPR lies in its ability to infer a posterior distribution over functions based on the data. This distribution provides a point estimate $\hat{y}^{\ast}$ at any given time $t$ and a measure of uncertainty as a confidence interval. This predictive distribution is central to prognostics as it naturally quantifies the uncertainty of the RUL prediction. In the application stage of prediction, when a new $\mathbb{X} ^{\ast } $ sampled from historical data is observed, temporal learning processes it into $\mathbf{h} ^{\ast } $. The joint prior Gaussian distribution of the training RUL $\mathbf{y}$ and  $y^{\ast }$ is obtained as follows,
\begin{eqnarray}
\begin{bmatrix}\mathbf{y}\\y^{\ast }\end{bmatrix}\sim N\left(\begin{bmatrix}\mathbf{\mu(\mathbf{h})}\\\mathbf{\mu(\mathbf{h} ^{\ast })}\end{bmatrix},\begin{bmatrix}\mathbf{K}+\delta^2I_m&\mathbf{K}(\mathbf{h} ^{\ast })\\\mathbf{K}(\mathbf{h} ^{\ast })^\intercal &k(\mathbf{h} ^{\ast },\mathbf{h} ^{\ast })\end{bmatrix}\right). \label{eq:quadratic10}   
\end{eqnarray}
The posterior mean function and covariance functions are given by
\begin{eqnarray}
\mu_m(\mathbf{h} ^{\ast })=\mathbf{K}(\mathbf{h} ^{\ast })^\intercal(\mathbf{K}+\delta^2I_m)^{-1}(\mathbf{y}-\mathbf{\mu(\mathbf{h})})+\mu(\mathbf{h} ^{\ast }),\label{eq:quadratic11}\\
k_m(\mathbf{h} ^{\ast })=k (\mathbf{h} ^{\ast },\mathbf{h} ^{\ast })-\mathbf{K}(\mathbf{h} ^{\ast })^\intercal(\mathbf{K}+\delta^2I_m)^{-1}\mathbf{K}(\mathbf{h} ^{\ast }),\label{eq:quadratic12}
\end{eqnarray}
where $\mathbf{K}(\mathbf{h} ^{\ast })=(k(\mathbf{h} ^{\ast },\mathbf{h}_1),\ldots,k(\mathbf{h} ^{\ast },\mathbf{h}_m))^\intercal $, the mean of RUL can be predicted using $\mu_m(\mathbf{h} ^{\ast })$, assuming a $GP(0,\mathbf{K})$ prior. The prediction is obtained as 
\begin{eqnarray}
{\hat{y}^{\ast}}= \mathbf{K}(\mathbf{h} ^{\ast })^\intercal(\mathbf{K}+\delta^2I_m)^{-1}\mathbf{y}. 
\end{eqnarray}

It is important to acknowledge that no prediction can be completely accurate, and prediction errors cannot be eliminated. To address this issue, we propose a straightforward approach providing a $1 - \alpha$ prediction interval such that 
\begin{eqnarray}
\Pr\{\hat{y}^{\ast}\in\left [   y ^{L}  , y ^{U} \right ]\} = 1-\alpha,
\end{eqnarray}
where $y ^{L} = \hat{y}^{\ast} - z_{\alpha/2}*\sqrt{k_m(\mathbf{h} ^{\ast })}$, $y ^{U} = \hat{y}^{\ast} + z_{\alpha/2}*\sqrt{k_m(\mathbf{h} ^{\ast })}$, and $z_{\alpha/2}$ represents the quantile of the corresponding standard normal distribution.Generally, when $\alpha =0.05$, the lower bound $y ^{L}$ is $\hat{y}^{\ast} - 1.96*\sqrt{k_m(\mathbf{h} ^{\ast })}$ and the upper bound $y ^{U}$ is $\hat{y}^{\ast} + 1.96*\sqrt{k_m(\mathbf{h} ^{\ast })}$.
This predictive distribution is central to prognostics as it naturally quantifies the uncertainty of the RUL prediction.  

In summary, GPR offers a probabilistic framework for RUL prediction that effectively accommodates the temporal dynamics inherent in prognostic processes. Additionally, it systematically quantifies uncertainty. The flexibility of modified GPR to analyze contributions from individual sensors further enhances its applicability for fault diagnosis and prognostics in complex mechanical systems.

\subsection*{Importance analysis}

A feature importance analysis is integrated into the model to determine which sensors most significantly influence the RUL. This analysis is based on evaluating the impact of each feature, derived from sensor data, on the RUL predictions. The feature importance assessment begins by inputting the hidden state $\mathbf{h}$, defined by equation (\ref{eq:hidden state}), into the analysis component. This hidden state encapsulates the relevant temporal features extracted during the data preprocessing and temporal feature extraction stages. The output of this process is a set of feature importance scores $\left \{ \lambda  ^{1} ,\lambda ^{2} ,...,\lambda ^{M} \right \} $, where each $\lambda$ corresponds to a sensor feature, and a higher $\lambda$ value indicates a more significant contribution to the machine's degradation and, thus, the RUL prediction.

The model evaluates feature importance by systematically altering the input features and measuring the corresponding changes in prediction accuracy. Specifically, it compares the accuracy of the full model against a version where each feature is individually permuted. Permutation involves randomly shuffling a given feature's values while keeping the others intact, disrupting its relationship with the output. By observing the decline in accuracy after this permutation, the model quantifies the importance of that specific feature. If the prediction accuracy significantly decreases, it implies that the feature in question plays a critical role in predicting RUL. Conversely, the feature is deemed less necessary if there is little to change.

This approach allows for an interpretability layer in the model, making it possible to identify which sensors are most likely to influence the degradation process. By ranking the features based on their importance scores, engineers can gain insights into which sensor data are the primary contributors to the machine's health decline and may indicate fault-prone areas in the system. This information is invaluable for maintenance planning, as it allows engineers to focus on the most critical sensor readings and take preventative actions based on the insights from the feature importance analysis.

The methodology of feature evaluation is inspired by the way decision trees assess the relevance of each input feature in complex models~\cite{breiman2017classification,li2018random}. The ranking produced by this process directly supports more effective and interpretable predictions in the context of fault diagnosis and predictive maintenance.

\section*{Experimental setup}

\subsection*{Dataset description}

We show the efficacy of the suggested approach using the C-MAPSS dataset as a benchmark. 'Commercial Modular Aero-Propulsion System Simulation', or C-MAPSS, is a NASA tool that simulates extensive commercial turbofan engine data using Matlab Simulink. C-MAPSS dataset is created using the C-MAPSS simulator~\cite{frederick2007user}. Four subsets exist, from FD001 to FD004, within the dataset~\cite{shi2024dual}. Twenty-one sensor measurements are gathered at each observation time, providing comprehensive information on engine locations and operational conditions. In the training set, each engine initially operates normally but starts to deteriorate after a specific time.
Conversely, the test set comprises incomplete data, with time series terminating before the onset of engine degradation~\cite{yan2016multiple}. The aim is to estimate each engine's remaining operable cycle count. The complete introduction of the dataset can be found in Supplementary Material under S1.

\subsection*{Configuration setting}

The architecture is implemented using Python 3.7 and the PyTorch 1.13.1 (GPU version) framework. The hardware configuration includes an Intel(R) Xeon(R) Platinum 8380 CPU, eight RTX 3090 GPUs, and 500 GB of RAM. 

\subsection*{Evaluation  setting}

To evaluate the model's performance, we utilize three functions. The first is the root mean square error (RMSE)~\cite{saxena2008damage}. Define $d=RUL _{predict} - RUL_{true} $, representing the difference between the predicted and true RUL values. RMSE  is calculated using
\begin{eqnarray}
 \mathrm{R M S E}=\left (   \frac{1}{M} \sum_{i=1}^{M} d_{i}{ }^{2}\right )^{1/2} , 
 \label{rmse}
\end{eqnarray}
where $M$ denotes the number of engines. The lower the RMSE score, the more accurate the interval prediction.

Two additional functions, average coverage interval length and coverage probability, are utilized to assess the model's interval prediction performance. Normalized averaged width (NAW) measures the average width of the constructed predicted intervals as a percentage of the underlying target range. The definition of NAW is provided as follows,
\begin{eqnarray}
\mathrm{NAW}=\frac1{RN}\sum_{j=1}^N\left(y_j^U-y_j^L\right),    
\label{naw}
\end{eqnarray}
where $R$ represents the range of the target variable throughout the forecast period. Predicted intervals with lower NAW values are considered more effective.

The coverage width-based criterion (CWC) provides a comprehensive evaluation score based on coverage probability and NAW. The fundamental concept behind CWC is that the score should be high irrespective of interval width if coverage probability is below the nominal confidence level. At the same time, NAW becomes the dominant factor if coverage probability exceeds this level. The definition of CWC  is
\begin{eqnarray}
\mathrm{CWC}=\mathrm{NAW}\times \exp(\frac{1-coverage}{\alpha } ),   
\label{cwc}
\end{eqnarray}
where $coverage = \frac1N\sum_{j=1}^NC_j$, where $C_{j}=1 $ if $\mathrm{RUL}_j \in [y_j^U,y_j^L]$, and  $C_{j}=0 $ otherwise. A lower CWC score indicates a more effective interval prediction.

\section*{Results}

\subsection*{Feature learning analysis}

% \vspace{2.5cm}
Figure \ref{FD001 distribution} displays the kernel density estimates (KDE) for sub-dataset FD001, providing the probability density function (PDF) of the 14 features individually. Besides, we provide importance ranking underneath. Since the other three sub-datasets exhibit similar trends and patterns, the Supplementary Materials under S4 give the full results. 

\begin{figure}[ht]
\centering
\includegraphics[scale=0.9]{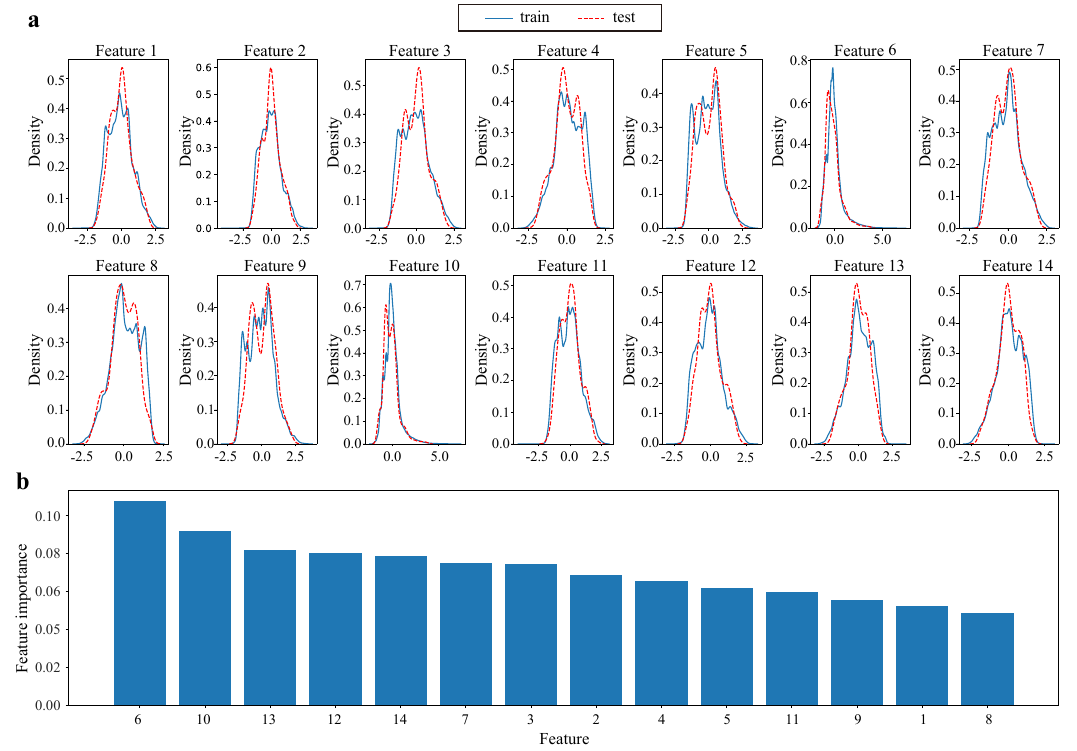}
\caption{Distribution and importance for sub-dataset FD001. Red dashed lines represent the density of testing datasets, and blue lines represent the density of training datasets. The lower figure shows the importance ranking of 14 features.}
\label{FD001 distribution}
\end{figure}

KDE provides a non-parametric approach to estimating a random variable's PDF. The default Gaussian kernel is used. The coincident density curves demonstrate strong consistency in the PDFs across the test and training datasets, thereby validating the use of GPR on the dataset. The histogram of importance shows that Feature 6 is the most important in FD001, demonstrating its robustness and significant predictive power across these datasets. Engineers can optimize the engine by focusing on the most influential variables, such as Features 6, 10, and 13.

\subsection*{Prognostic results analysis}

\begin{figure}[htbp]
\centering
\includegraphics[scale=0.8]{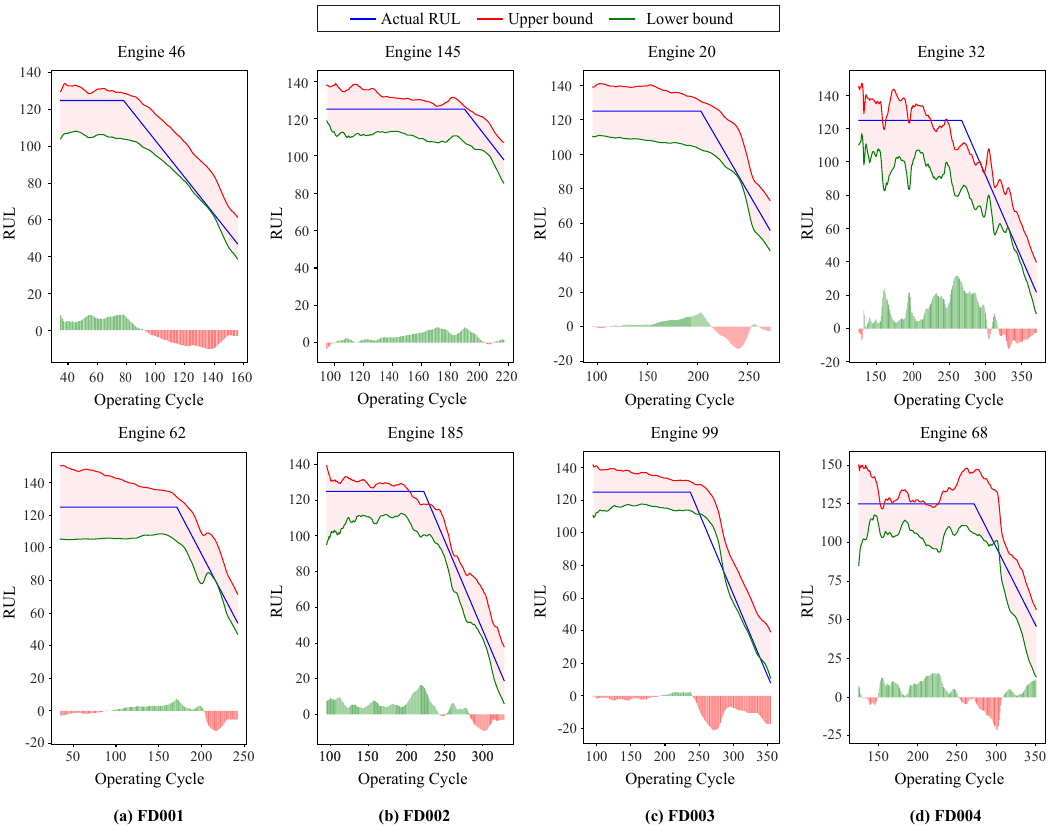}
\caption{RUL prognostic performances of our algorithm for the testing engine units in four sub-datasets. The blue polyline represents the actual RUL. The range between the red curve and the green curve represents the predicted interval. The red segment indicates that the predicted RUL exceeds the actual RUL, misleading engineers into believing the machine can still operate. Conversely, the green segment signifies that the predicted RUL is less than the actual RUL. Engine 46 and 62 belong to FD001. Engine 145 and 185 belong to FD002. Engine 20 and 99 belong to FD003. Engine 32 and 68 belong to FD004.}
\label{single}
\end{figure}

To further test the performance of our model, eight engines, two from each of the four test datasets, are randomly selected. The actual RUL curves and online RUL interval estimation results for the test set, at a 95\% confidence level, are presented in Fig. \ref{single}. The errors between ground truth and predicted points are plotted in Fig. \ref{single}. The lower segment shows that although the predicted RUL does not always match the actual RUL, it does not lead to poor judgment. The predicted intervals, composed of upper and lower prediction limits, appropriately cover the target values. As service time progresses, the upper and lower prediction limits closely follow the actual value changes. Although the predicted RUL sometimes exceeds the actual RUL, the predicted intervals consistently cover the real RUL, demonstrating the utility of the proposed method for constructing prediction intervals to capture fault trends.

\subsection*{Comparison with the state-of-the-art methods}
\begin{table}[ht]
\centering
\begin{tabular}{|l|l|l|l|l|}
\hline
Models&  FD001&FD002&FD003&FD004\\
\hline
D-LSTM~\cite{zheng2017long}, 2017&16.14 &24.49 &16.18&28.17\\
BS-LSTM~\cite{liao2018uncertainty}, 2018&14.89&26.86&15.11&27.11\\
BL-CNN~\cite{liu2019novel}, 2019 &13.18&19.09&13.75&20.97 \\
RF-LSTM~\cite{tang2020improvement}, 2020&16.87&23.91&17.89&25.49\\
CNN-LSTM~\cite{mo2021evolutionary}, 2021&\textbf{11.56}&17.67&12.98&20.19 \\
DA-Transformer~\cite{liu2022aircraft}, 2022&12.25&17.08&13.39&19.86 \\
IDMFFN~\cite{zhang2023integrated}, 2023&12.18&19.17&\textbf{11.89}&21.72 \\
        HRP(Ours)& 13.09& \textbf{12.33}&13.49& \textbf{19.65}\\
\hline
\end{tabular}
\caption{\label{Comparison study}Comparison of different models on RMSE.}
\end{table}

To provide a comprehensive evaluation of each model’s performance under various operating conditions and to facilitate a clear comparison of their strengths and limitations, the key metrics RMSE (\ref{rmse}), NAW (\ref{naw}), and CWC (\ref{cwc}) have been calculated for each algorithm. As shown in Table \ref{Comparison study}, our proposed method achieves commendable RMSE results across the four sub-datasets, outperforming most established methods in most cases. This highlights the robustness of our approach in accurately predicting RUL, especially under complex conditions. The results are particularly noteworthy for the FD002 subset, encompassing more challenging operational settings with diverse fault modes. While the performance of our model on sub-datasets FD001 and FD003 does not achieve the absolute best RMSE, it remains competitive and within the lower range of error rates compared to other models. This indicates that even in cases where our model does not achieve the top score, it still provides a highly reliable prediction with minimal deviation, ensuring robust and consistent results across varying conditions. This consistency is essential for practical applications in industrial prognostics, as it supports dependable predictions regardless of the specific operational scenario.
\begin{table}[ht]
\centering
\begin{tabular}{|l|l|l|l|l|l|l|l|l|}
\hline
~& FD001&~&FD002&~&FD003&~&FD004&~ \\
\hline
Models&NAW(\%)&CWC(\%)&NAW(\%)&CWC(\%)&NAW(\%)&CWC(\%)&NAW(\%)&CWC(\%)\\
\hline
LSTMBS~\cite{liao2018uncertainty},2018&~37.70&83.90&47.20&89.20&45.90&102.15&65.40&170.80\\
IESGP~\cite{liu2019multiple}, 2019&~54.00&59.68&55.70&137.00&44.50&93.27&49.10&248.11\\
AGCNN~\cite{liu2023uncertainty}, 2023&~48.06&97.00&59.29&98.46&44.26&97.00&63.48&\textbf{95.56}\\
HRP(Ours)&~\textbf{21.00}&\textbf{38.48}&\textbf{29.00}&\textbf{76.06}&\textbf{23.00}&\textbf{27.94}&\textbf{28.00}&468.43\\
\hline
\end{tabular}
\caption{\label{Interval comparison study}Comparison of different models on interval criterions.}
\end{table}

In addition to the RMSE analysis, an interval comparison for the parameters NAW and CWC is provided in Table \ref{Interval comparison study}. These metrics offer insights into the reliability and precision of the prediction intervals produced by each model. Across all four sub-datasets, our approach consistently reduces NAW values by at least 50.70\% on average compared to other models. This reduction demonstrates the model’s capability to provide tighter prediction intervals, which are desirable for practical applications where narrow intervals indicate higher confidence in the predictions. Furthermore, the CWC values, which combine interval width and coverage probability to evaluate the overall quality of prediction intervals, also significantly improve. On the first three sub-datasets, FD001, FD002, and FD003, our method achieves an average CWC reduction of at least 50.16\%, suggesting that the intervals become narrower and maintain appropriate coverage. This balance of narrow intervals with adequate coverage reflects high accuracy and reliability, essential for predicting RUL in smart manufacturing settings. Out of the four sub-datasets, our method consistently outperforms existing approaches, providing the most reliable and interpretable intervals. This suggests that our model holds considerable promise for effectively predicting faults in advance, contributing to robust maintenance planning and improved operational safety in industrial systems.

\section*{Conclusion}

This paper proposes an intelligent RUL interval prediction method based on a modified GPR network for aero-engines. We embedded a temporal feature extraction into regression, enhancing the accuracy and robustness of RUL interval predictions. As an effective regression method, GPR is modified to adapt to engineering properties in our network, which comprehensively learns the diversity of distributions between sensors and RUL. Results from experiments using the C-MAPSS dataset show that the proposed method significantly narrows the width of the 95\% confidence interval and enhances coverage accuracy, thereby aiding in the PHM tasks of detection, diagnostics, and prognostics. Additionally, by adaptively employing additional random forest regression, the influence of sensors on fault modes can be assessed, facilitating predictive maintenance decisions for the reliable operation of machinery components.

Future work will incorporate transfer learning across different but similar sub-datasets to enhance predictive outcomes. Addressing the challenge of exponentially increasing training times in Gaussian regression as data size grows presents another intriguing research direction.

\section*{Data availability}

The data that support the findings of this study are available from the corresponding author upon reasonable request. 

\bibliography{for_arxiv}

\section*{Acknowledgments}
This study was partially supported by Shanghai Municipal Science and Technology Major Project (No.2021SHZDZX0103). This study was also supported by (1) Shanghai Engineering Research Center of AI $\&$ Robotics, Fudan University, China, and (2) Engineering Research Center of AI $\&$ Robotics, Ministry of Education, China.

\section*{Author contributions statement}

Tian Niu conceived the methodology and experiment(s), conducted the investigation, and contributed to the original draft preparation and validation. Zijun Xu edited the manuscript, participated in the investigation, and validated the results. Heng Luo contributed to the editing of the manuscript, as well as the development of the methodology. Ziqing Zhou is the corresponding author, conceiving the methodology. All authors reviewed the manuscript. 

\section*{Competing interests}

The author(s) declare no competing interests. 

\section*{Additional information}

\textbf{Competing interests}  The authors declare no competing interests. 

\newpage

\title{Supplementary Materials}

\date{}

\maketitle

\section*{S1. C-MAPSS Dataset}

\begin{figure}
    \centering
    \includegraphics[width=\textwidth]{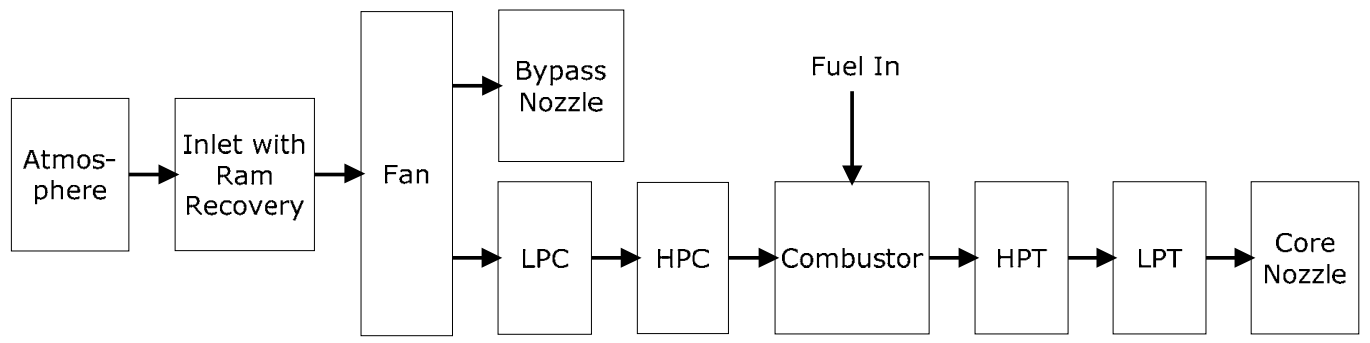}
    \caption{Subroutines of the engine simulation.}
    \label{fig:engine}
\end{figure}

The subroutines of the engine simulation with ducts and bleed omitted is shown in Fig. \ref{fig:engine}.

The dataset comprises full flight recordings sampled at a frequency of 1 Hz, capturing data from 30 engines across various flight condition parameters. Each flight spans approximately 90 minutes and includes seven unique flight conditions, including the climb to a 35,000-foot cruise altitude and the following descend to sea level. Temperature and pressure readings, health indicators, and flight conditions are among the metrics that are recorded for every flight.

\begin{table}[ht]
\centering
\begin{tabular}{|l|l|l|l|l|}
\hline
Dataset description&  FD001& FD002& FD003& FD004\\
\hline
Number of engines in the training set& 100& 260& 100& 249\\
Number of engines in the test set& 100& 259& 100& 248\\
Operating conditions& 1& 6& 1& 6\\
Fault modes& 1& 1& 2& 2\\
\hline
\end{tabular}
\caption{\label{CMAPSS description}Description of the C-MAPSS dataset.}
\end{table}

Table \ref{CMAPSS description} illustrates that there are four subsets within the dataset. Sub-datasets FD001 and FD002 feature a single fault mode, which is HPC (High Pressure Compressor) Degradation. The remaining two sub-datasets include an additional fault mode, Fan Degradation.

\begin{table}[ht]
\centering
\begin{tabular}{|l|l|l|}
\hline
Symbol &Description &Units    \\
\hline
T2 &Total temperature at fan inlet& °R \\
T24 &Total temperature at LPC outlet &°R \\
T30 &Total temperature at HPC outlet& °R \\
T50 &Total temperature at LPT outlet &°R\\
P2 &Pressure at fan inlet &psia \\
P15 &Total pressure in bypass-duct &psia \\
P30 &Total pressure at HPC outlet& psia\\
Nf &Physical fan speed &r/min \\
Nc &Physical core speed& r/min\\
epr& Engine pressure ratio(P50/P2)& - \\
Ps30& Static pressureat HPC outlet& psia \\
phi &Ratio of fuel flow to Ps30 &pps/psi\\
NRf &Corrected fan speed &r/min \\
NRc &Corrected core speed &r/min\\
BPR &Bypass Ratio& - \\
farB& Burner fuel-air ratio& -\\
htBleed& Bleed Enthalpy & -\\
Nf\_dmd &Demanded fan speed &r/min\\
PCNfR\_dmd& Demanded corrected fan speed &r/min \\
W31& HPT coolant bleed& lbm/s \\
W32& LPT coolant bleed& lbm/s \\
\hline
\end{tabular}
\caption{\label{CMAPSS 21 sensors}Detailed description of the 21 sensors.}
\end{table}

Each dataset is further separated into training and test sets and comprises 21 sensor signals, with detailed information about the sensors provided in Table \ref{CMAPSS 21 sensors}. 

\begin{table}[ht]
\centering
\begin{tabular}{|l|l|l|l|}
\hline
Operation& Altitude(Kft) &  Machnumber & TRA\\
\hline
1& 35& 0.8400& 100  \\
2 &42 &0.8408 &100 \\
3 &25& 0.6218& 60 \\
4 &25 &0.7002 &100\\
5 &20 &0.2516 &100\\
6 &10& 0.7002 &100 \\
\hline
\end{tabular}
\caption{\label{CMAPSS operational conditions}Detailed description of different operational conditions.}
\end{table}

The degradation dataset encompasses six operational conditions featuring various combinations of Altitude, Throttle Resolver Angle (TRA), and Mach number, detailed in Table \ref{CMAPSS operational conditions}. Notably, the changes in operational conditions for each unit occur in a relatively random pattern, and the operational conditions vary significantly across different units. 

\section*{S2. Data preprocessing}

\subsubsection*{Sensor data selection}

Each sub-dataset in the C-MAPSS dataset comprises 21 sensor signals. To enhance the accuracy of RUL predictions while reducing computational complexity, strategic sensor selection is necessary. Because, sensors 1, 5, 6, 10, 16, 18 and 19 stay consistent, only the remaining 14 sensor signals are retained as they provide critical information for effective prognostics.

The parameters are updated by the backpropagation technique. Since RUL is a supervised learning method, it needs to be given true throughout training. The piecewise linear degradation model are the most often used models to characterize RUL. Here, the time for the change point is set at 125, thus

$  RUL_{final}=\begin{cases}
  & 125~~~~~~~~~\text{ if } RUL_{real}\ge 125 \\
  & RUL_{real}~~\text{ if } RUL_{real}<   125
\end{cases}
$.

\subsubsection*{Sensor data normalization}

Before the sensor readings for the chosen sensor sets are entered into the prediction network, we standardize them using z-score normalization. This normalization transforms the original time series $x_{i}^{j}$, corresponding to the $j$-th sensor of the $i$-th sample, as formula
\begin{equation}
   x_{i}^{j\prime}=\frac{x_{i}^{j}-\mu^{j}}{\sigma^{j}},
\end{equation}
where $\mu^{j}$ and $\sigma^{j}$ represent the mean value and the standard deviation deviation, respectively, for the $j$-th sensor readings across all instances in each dataset. The raw data show significant fluctuations and no discernible pattern pertaining to the engine's life cycle. The data show a clear growing trend within an acceptable range after normalization, demonstrating the need to apply the z-score normalization procedure to the raw data prior to using them in prediction models.

\subsubsection*{Sensor data smoothing}

There is a lot of noise in the initial sensor output, which could lead to the model learning unrelated properties. In order to preserve the trend of sensor signals while decreasing oscillation and eliminating unnecessary features, we adopt exponential smoothing. The exponential smoothing equation is governed by a smoothing parameter $s$, which controls the smoothness of the resulting smoothed series. A larger $s$ value results in a smoother series. The formula for calculating the smoothed value at time $t$ is
\begin{eqnarray}
   \tilde{\mathbb{X} } _{t} = \beta  \mathbb{X}  _{t}+  \left ( 1- \beta   \right ) \tilde{\mathbb{X} } _{t- 1}, 
\end{eqnarray}
 where $\tilde{\mathbb{X} } _{t}$ represents the smoothed value at time $t$, $\mathbb{X}  _{t}$ is the observed value at time $t$, and $\beta $ is the smoothing parameter, ranging between 0 and 1. This parameter determines the weight of the current observation relative to the previous smoothed value. Typically, $\beta $ is calculated using the formula $\beta  = \frac{2}{1+s}$, where $s$ is a smoothing coefficient. 

\subsubsection*{Sliding window processing}

A transient approximation of the real values of the time series data is the sliding window. The window and segment sizes are increased until a less inaccurate approximation is obtained. Following the first segment's selection, the next section is picked, commencing at the conclusion of the previous segment. Until all time series data have been segmented, the procedure is repeated. 

\begin{figure}
    \centering
    \includegraphics[width=\textwidth]{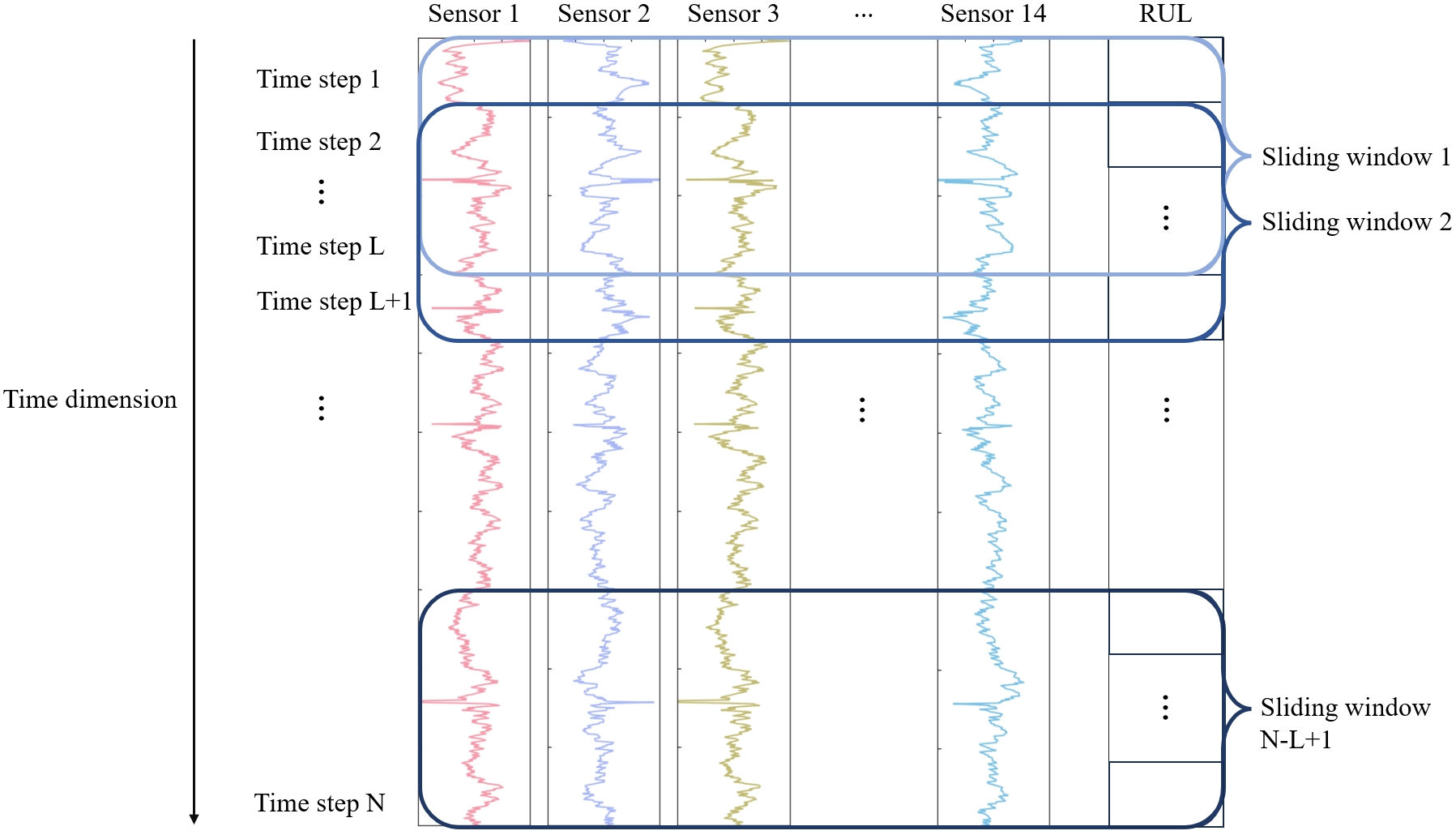}
    \caption{Illustration of the sliding window data segmentation. Sliding window process separates whole time dimension into equal length time window.}
    \label{sliding window}
\end{figure}

The sliding window process is shown in Fig. \ref{sliding window} with $window~size=L$ and $sliding~stride=1$. In order to forecast the final time step RUL, the sliding window gathers past time series data. In actuality, raw data is used to calculate window size. Performance is improved by longer time periods since they catch more important information. Longer time periods, however, might make the model more complex and limit its application. 
\begin{table}[ht]
\centering
\begin{tabular}{|l|l|l|l|l|}
\hline
~&  FD001& FD002 &FD003&FD004 \\
\hline
Time window size & 25 & 20 &30& 15\\
Min life cycle in training datasets  & 128& 128& 145 & 128\\
Number of training time windows & 18231 & 48819 & 21820 & 57763 \\
Min life cycle in test datasets & 31 & 21 & 38 &19\\
Number of testing time windows & 10696  & 29070 & 13696 & 37742 \\
\hline
\end{tabular}
\caption{\label{CMAPSS window}Details of time window setting for each sub-dataset.}
\end{table}
The appropriate window length has been selected as detailed in Table\ref{CMAPSS window}. 

\section*{S3. Temporal feature extraction}

The following set of mathematical expressions outlines how these temporal features are computed at each time step $t$, where $t=1,2,..,T$ , 
\begin{eqnarray}
f_t=\sigma(W_\mathrm{f}\cdot[h_{t-1},x_t]+b_\mathrm{f}),\label{eq:quadratic1}\\
i_t=\sigma(W_i\cdot[h_{t-1},x_t]+b_i),\label{eq:quadratic2}\\
\tilde{C}_t=\tanh(W_C\cdot[h_{t-1},x_t]+b_C)W_C,\label{eq:quadratic3}\\
C_t=f_t\cdot C_{t-1}+i_t\cdot\tilde{C}_t,\label{eq:quadratic4}\\
o_t=\sigma(W_\mathrm{o}\cdot[h_{t-1},x_t]+b_\mathrm{o}),\label{eq:quadratic5}\\
h_t=o_t*\tanh(C_t).\label{eq:quadratic6}
\end{eqnarray}

The variables $f_t$, $i_t$, $o_t$ represent the forgetting gate, input gate, and output gate, respectively. $\sigma$ in equation (\ref{eq:quadratic1}), equation (\ref{eq:quadratic2}) and equation (\ref{eq:quadratic6}) denotes the sigmoid function, and $\tanh$ in equation (\ref{eq:quadratic3}) is the hyperbolic tangent function. The $\sigma$ and $\tanh$ functions introduce nonlinear characteristics into the temporal feature extraction network. The variable $x_t$ represents the current input, $\tilde{C}_t$ is the cell state of the current input, and $C_t$ is the cell state at the current moment, used to store memory. Equation (\ref{eq:quadratic4}) illustrates the mechanism for updating $C_t$ by incorporating the forgetting and addition processes. $h_{t-1}$ denotes the hidden state. The weights $W_\mathrm{f}$, $W_\mathrm{i}$, $W_\mathrm{C}$, $W_\mathrm{o}$ and biases $b_\mathrm{f}$, $b_\mathrm{i}$, $b_\mathrm{C}$, $b_\mathrm{o}$ are updated during training. 

\section*{S4. Distribution and Importance}

\begin{figure}[htbp]
\centering
\includegraphics[scale=0.9]{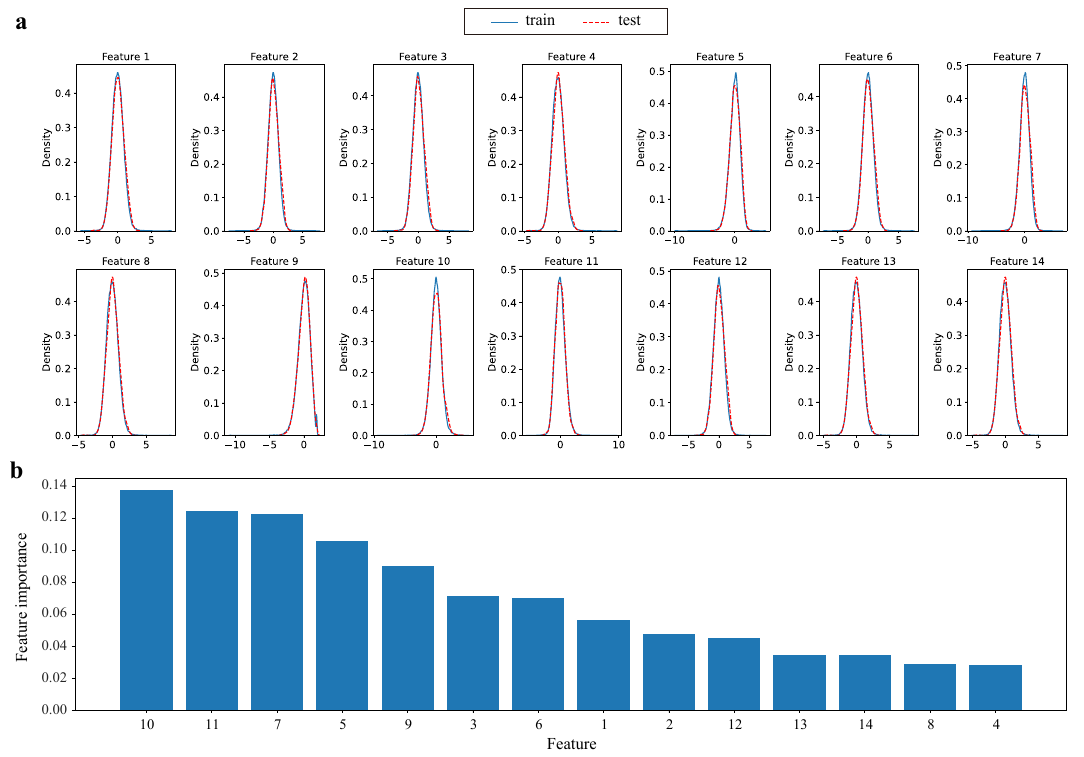}
\caption{Distribution and importance for sub-dataset FD002. Red dashed lines represent the density of testing datasets, and blue lines represent the density of training datasets. The lower figure shows the importance ranking of 14 features.}
\label{FD001 distribution}
\end{figure}
\begin{figure}[htbp]
\centering
\includegraphics[scale=0.9]{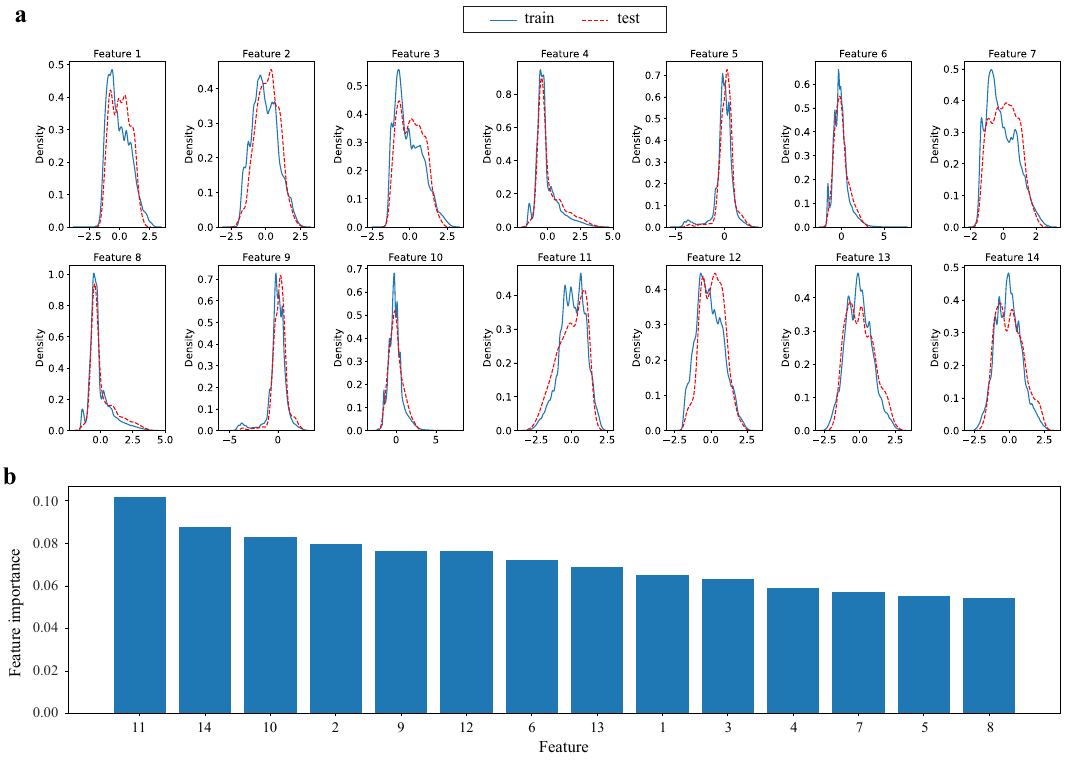}
\caption{Distribution and importance for sub-dataset FD003. Red dashed lines represent the density of testing datasets, and blue lines represent the density of training datasets. The lower figure shows the importance ranking of 14 features.}
\label{FD001 distribution}
\end{figure}
\begin{figure}[htbp]
\centering
\includegraphics[scale=0.9]{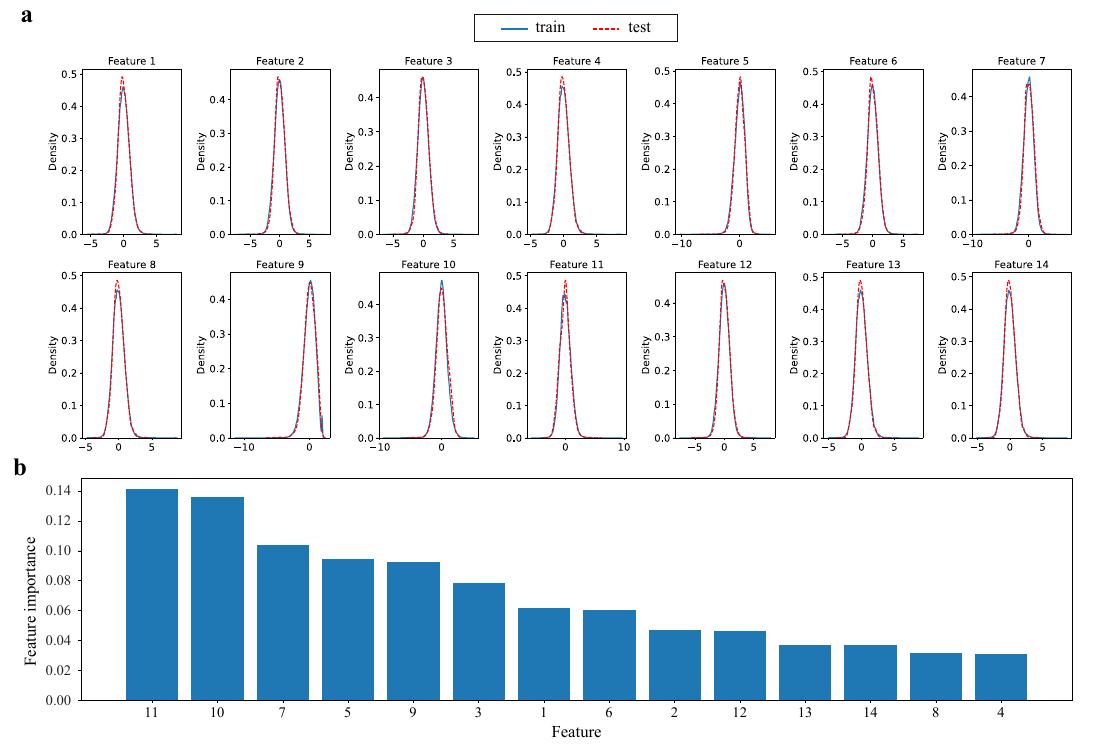}
\caption{Distribution and importance for sub-dataset FD004. Red dashed lines represent the density of testing datasets, and blue lines represent the density of training datasets. The lower figure shows the importance ranking of 14 features.}
\label{FD001 distribution}
\end{figure}

\end{document}